\documentclass[journal]{IEEEtran}
\IEEEoverridecommandlockouts
\usepackage{graphicx}
\usepackage{amsmath}
\usepackage{hyperref}
\usepackage{cite}
\usepackage{booktabs}
\usepackage{pifont}
\usepackage{colortbl}
\usepackage{xcolor}

\title{YOLOv1 to YOLOv11: A Comprehensive Survey of Real-Time Object Detection Innovations and Challenges}

\author{\IEEEauthorblockN{ Manikanta Kotthapalli,}
\and
\IEEEauthorblockN{ Deepika Ravipati, }
\and
\IEEEauthorblockN{ Reshma Bhatia}
}

\begin{document}

\maketitle

\begin{abstract}
Over the past decade, object detection has advanced significantly, with the YOLO (You Only Look Once) family of models transforming the landscape of real-time vision applications through unified, end-to-end detection frameworks. From YOLOv1’s pioneering regression-based detection to the latest YOLOv9, each version has systematically enhanced the balance between speed, accuracy, and deployment efficiency through continuous architectural and algorithmic advancements.. Beyond core object detection, modern YOLO architectures have expanded to support tasks such as instance segmentation, pose estimation, object tracking, and domain-specific applications including medical imaging and industrial automation. This paper offers a comprehensive review of the YOLO family, highlighting architectural innovations, performance benchmarks, extended capabilities, and real-world use cases. We critically analyze the evolution of YOLO models and discuss emerging research directions that extend their impact across diverse computer vision domains.
\end{abstract}


\section{Introduction}

Object detection is a cornerstone task in computer vision, involving both the identification and localization of objects within static images or dynamic video streams. Over the past decade, deep learning has transformed this field, delivering significant gains in detection accuracy, robustness, and deployment efficiency. Among the most influential developments is the YOLO (You Only Look Once) family of models, which redefined real-time object detection by introducing a unified, end-to-end architecture capable of delivering high-speed predictions without sacrificing accuracy.

Traditional object detectors such as R-CNN~\cite{girshick2014rich}, Fast R-CNN~\cite{girshick2015fast}, and Faster R-CNN~\cite{ren2015faster} adopt a two-stage approach: first generating region proposals and then classifying and refining these regions. While these methods have demonstrated high accuracy, their sequential architecture incurs significant computational overhead, making them unsuitable for latency-sensitive applications such as autonomous driving, robotics, and real-time surveillance.

YOLO disrupted this paradigm by reframing object detection as a single-stage regression problem. It predicts bounding boxes and class probabilities directly from full images in a single forward pass~\cite{redmon2016you}. This shift enabled orders-of-magnitude speed improvements while maintaining competitive accuracy. Since its inception with YOLOv1, the YOLO family has evolved rapidly, with each version introducing architectural innovations, optimization techniques, and deployment improvements. More recent iterations, such as YOLOv8 and YOLOv9, have extended the framework’s applicability to tasks including segmentation, pose estimation, and edge-device deployment.

This paper presents a comprehensive review of the YOLO series, from YOLOv1 through YOLOv9, with brief discussions of emerging models like YOLOv10 and YOLOv11. We examine the key architectural and algorithmic innovations across YOLO versions, from YOLOv1 through YOLOv9, highlighting how each iteration improved the trade-off between detection speed and accuracy. Performance benchmarks such as mAP and FPS are contextualized using results on standard datasets like PASCAL VOC and MS COCO. We further discuss the deployment characteristics of different YOLO variants across edge and server environments, noting their suitability for tasks such as small object detection, real-time inference, and multi-task learning (e.g., segmentation and pose estimation). Finally, we identify open challenges—such as training stability in anchor-free variants, robustness under domain shift, and interpretability—and propose future research directions that could improve the usability, efficiency, and adaptability of YOLO models in diverse real-world applications.

\section{Background on Object Detection}

Object detection is a fundamental task in computer vision that involves both identifying object classes and localizing them within an image. Early approaches such as Deformable Part Models (DPM) and Selective Search relied heavily on handcrafted features and exhaustive region proposal mechanisms~\cite{felzenszwalb2008discriminatively,uijlings2013selective}. While these methods laid the groundwork for detection frameworks, they were often limited by high computational cost and suboptimal performance on complex scenes.

The advent of deep learning revolutionized object detection by enabling end-to-end trainable architectures capable of learning robust feature representations directly from data. Early deep learning-based detectors adopted a two-stage approach: the first stage generates candidate object regions, and the second stage classifies and refines these proposals. Representative models include R-CNN~\cite{girshick2014rich}, Fast R-CNN~\cite{girshick2015fast}, and Faster R-CNN~\cite{ren2015faster}. In particular, Faster R-CNN introduced the concept of Region Proposal Networks (RPNs), integrating proposal generation into the deep learning pipeline and significantly improving efficiency and accuracy.

Despite their effectiveness, two-stage detectors tend to be computationally intensive, with inference times unsuitable for latency-sensitive applications such as robotics, autonomous driving, or real-time video analytics. To address this limitation, one-stage detectors were introduced. These models eliminate the explicit proposal generation phase and directly predict object classes and bounding boxes over dense spatial grids. Notable examples include the Single Shot MultiBox Detector (SSD)~\cite{liu2016ssd} and the original YOLO (You Only Look Once) model~\cite{redmon2016you}.

YOLO fundamentally reframed object detection as a regression problem, simultaneously predicting bounding boxes and class probabilities in a single forward pass through the network. This design enabled unprecedented inference speed, making YOLO a popular choice for real-time applications. However, early YOLO versions struggled with detection of small objects and exhibited lower localization precision compared to their two-stage counterparts.

Subsequent YOLO iterations have addressed these shortcomings through a series of architectural enhancements, including multi-scale feature fusion, advanced backbone networks, decoupled detection heads, and improved training strategies. Today, YOLO models constitute one of the most prominent families of one-stage object detectors, offering a compelling balance between speed, accuracy, and deployment flexibility.

In the following sections, we present a chronological review of the YOLO family from YOLOv1 through YOLOv9, highlighting the architectural innovations, performance improvements, and design philosophies that have driven their evolution.

\section{Taxonomy of YOLO Innovations}\label{sec:taxonomy}

To clarify the evolution of the YOLO family beyond sequential versioning, we categorize innovations along five orthogonal axes:

\begin{enumerate}
    \item \textbf{Backbone}: Feature extraction architecture.
    \item \textbf{Neck}: Feature fusion and multi-scale aggregation.
    \item \textbf{Detection Head}: Prediction structure and decoupling.
    \item \textbf{Loss and Assignment}: Optimization objectives and label matching.
    \item \textbf{Training Strategies}: Data augmentation, regularization, and architectural tricks.
\end{enumerate}

This taxonomy highlights the incremental and independent advances across components, facilitating modular understanding and design reuse.

\subsection{Backbone Evolution}
\noindent
Early YOLO versions (v1–v2) relied on classification-derived backbones such as GoogLeNet and Darknet-19. YOLOv3 introduced Darknet-53, a deeper, residual-style backbone. Subsequent versions adopted cross-stage partial connections (CSPDarknet in YOLOv4/v5) and, more recently, GELAN in YOLOv9 to improve both capacity and efficiency while maintaining real-time constraints.

\subsection{Neck and Feature Fusion}
\noindent
Multi-scale feature fusion became explicit in YOLOv3 and matured in later versions. YOLOv4 incorporated PANet to propagate context from shallow to deep layers. YOLOv7 introduced RepPAN, and YOLOv9 integrated GELAN-FPN, significantly enhancing localization of small and occluded objects through richer feature pyramids.

\subsection{Detection Head}
\noindent
Until YOLOv5, detection heads were largely coupled—jointly predicting class scores and bounding box regressions. YOLOv6 introduced a fully decoupled head architecture, allowing specialization of features. YOLOv8 transitioned to anchor-free prediction, while YOLOv9 merged decoupling with programmable gradient routing for dynamic task emphasis.

\subsection{Loss and Assignment}
\noindent
YOLOv1–v2 employed squared-error loss for bounding boxes. YOLOv3 onward adopted IoU-based losses (GIoU, CIoU), and YOLOv7–v9 adopted Distribution Focal Loss (DFL). In label assignment, the shift from greedy matching to SimOTA and OTA methods (e.g., in YOLOv6–v9) stabilized training and improved convergence in dense scenes.

\begin{figure}[!t]
\centering
\includegraphics[width=0.6\linewidth]{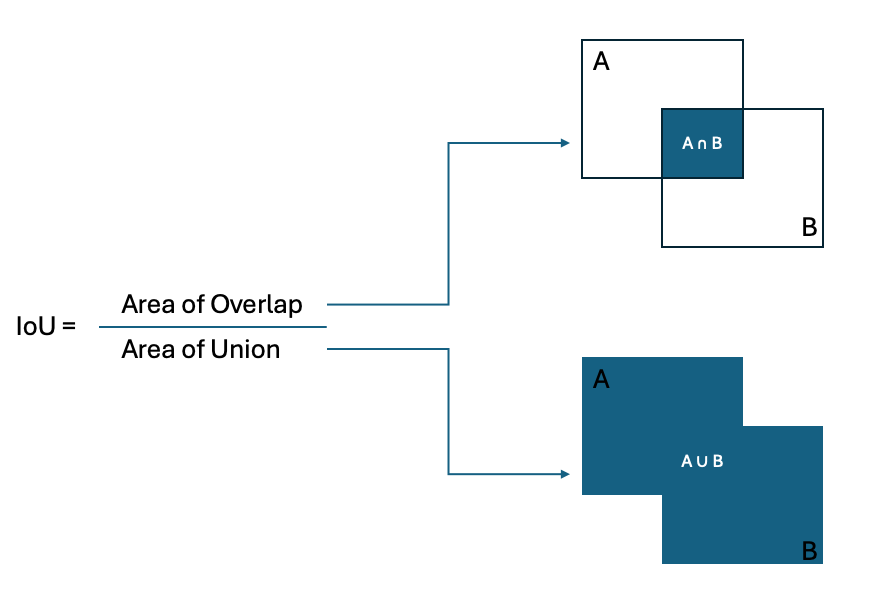} 
\caption{Intersection over Union (IoU) is computed as the ratio of the overlapping area to the union area between predicted and ground-truth boxes.}
\label{fig:iou}
\end{figure}

Intersection over Union (IoU), the standard metric for evaluating detection overlap, is illustrated in Fig.~\ref{fig:iou}, where IoU is defined as the ratio of the area of overlap to the area of union between predicted and ground-truth boxes.

\subsection{Training Strategies}
\noindent
Training pipelines evolved dramatically: YOLOv2 introduced multi-scale inputs; YOLOv4 added Mosaic and CutMix augmentation; YOLOv5 employed auto-anchor selection and hyperparameter evolution. YOLOv6/7 added re-parameterized convolutions and EMA; YOLOv8/9 integrated self-adversarial training and programmable assignment to balance precision and recall.

\begin{table*}[!t]
  \centering
  \caption{Innovation matrix across YOLO generations 
           (\ding{108}=major contribution, 
            \ding{117}=minor/first use, 
            --=not present).}
  \label{tab:yolo-taxonomy}
  \renewcommand{\arraystretch}{1.15}
  \setlength{\tabcolsep}{6pt}
  \rowcolors{3}{white}{gray!3}
  \begin{tabular}{lccccccccc}
    \toprule
    Category & v1 & v2 & v3 & v4 & v5 & v6 & v7 & v8 & v9 \\ \midrule
    Backbone              & \ding{108} & \ding{108} & \ding{108} & \ding{108} & \ding{108} & \ding{108} & \ding{108} & \ding{108} & \ding{108} \\
    Neck / FPN            & -- & -- & \ding{117} & \ding{108} & \ding{108} & \ding{108} & \ding{108} & \ding{117} & \ding{108} \\
    Head (decoupled)      & -- & -- & -- & -- & -- & \ding{108} & \ding{108} & \ding{108} & \ding{108} \\
    Anchor-free           & -- & -- & -- & -- & -- & \ding{117} & -- & \ding{108} & \ding{108} \\
    Loss (IoU/DFL)        & L2 & L2+IoU & BCE & CIoU & CIoU & vFL & vFL & DFL & DFLv2 \\
    Training Tricks       & -- & Multi-scale & Multi-scale & Mosaic/CutMix & Auto-Aug & EMA/Copy-Paste & RepConv & EMA+AF & SimOTA+EMA \\
    \bottomrule
  \end{tabular}
\end{table*}

\begin{figure}[t]
  \centering
  \includegraphics[width=0.9\columnwidth]{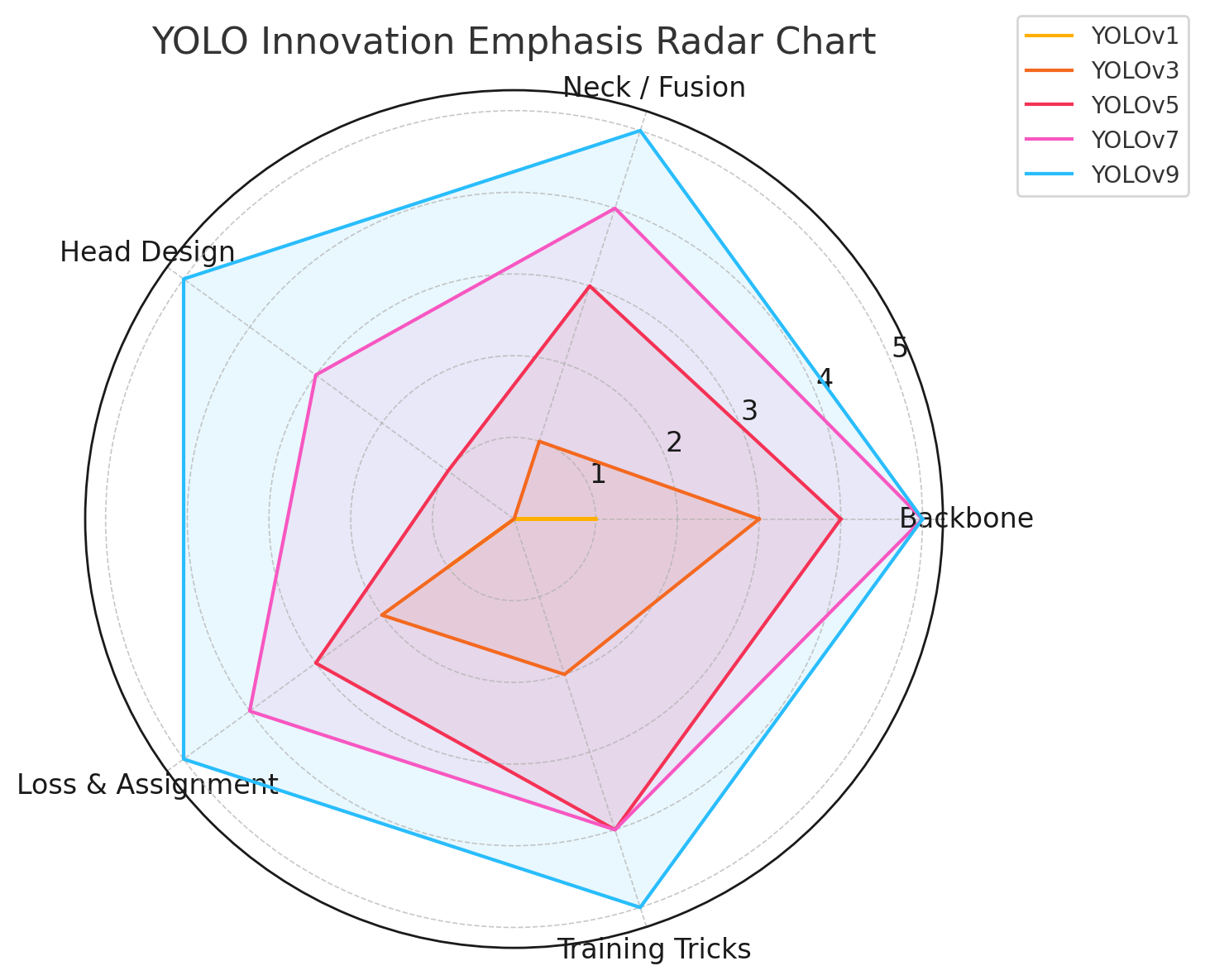}
  \caption{Radar plot of the relative emphasis of each YOLO version on the five innovation axes.}
  \label{fig:taxonomy-radar}
\end{figure}

A visual comparison of YOLO versions across five innovation axes is shown in Fig.~\ref{fig:taxonomy-radar}, highlighting the relative maturity of each component.

\section{YOLOv1: Unified Real-Time Object Detection}

YOLOv1 (You Only Look Once)~\cite{redmon2016you} marked a paradigm shift in object detection by framing the task as a single regression problem. Instead of relying on region proposals and separate classification stages, YOLOv1 introduced a fully end-to-end, single-stage network that directly predicts bounding box coordinates and class probabilities from an input image in a single forward pass. This unification enabled real-time inference speeds without the need for post-processing stages traditionally seen in two-stage detectors.

The YOLOv1 architecture is based on a custom convolutional neural network inspired by GoogLeNet~\cite{szegedy2015going}, optimized for speed and efficiency. It consists of 24 convolutional layers followed by two fully connected layers. The input image is divided into an $S \times S$ grid (with $S=7$), where each grid cell is responsible for detecting objects whose centers fall within the cell. Each grid cell predicts $B=2$ bounding boxes, associated confidence scores, and $C$ class probabilities.

YOLOv1 uses a multi-part loss function composed of three components: localization loss (for bounding box coordinates), confidence loss (for objectness prediction), and classification loss (for class probabilities), all trained jointly using standard backpropagation. The model achieved 45 FPS on the PASCAL VOC 2007 dataset, demonstrating that real-time object detection was achievable on standard hardware.

\subsection{Key Contributions}

\begin{itemize}
    \item \textbf{Unified Detection Framework:} YOLOv1 introduced a single-stage, fully end-to-end architecture that treats detection as a regression problem, simplifying the pipeline and enabling joint optimization.
    \item \textbf{High Inference Speed:} Achieved 45 FPS with a mean Average Precision (mAP) of 63.4\% on PASCAL VOC 2007, significantly outperforming prior detectors in terms of speed.
    \item \textbf{Global Context Reasoning:} By processing the entire image at once, YOLOv1 incorporated global contextual information, helping to reduce background false positives.
\end{itemize}

\subsection{Limitations}

\begin{itemize}
    \item \textbf{Localization Precision:} The coarse grid-based prediction leads to poor localization performance for small or densely packed objects.
    \item \textbf{Prediction Rigidity:} Each grid cell predicts a fixed number of bounding boxes, limiting the model’s ability to detect multiple adjacent objects.
    \item \textbf{Lower Accuracy Compared to Two-Stage Models:} Despite its speed, YOLOv1 underperforms in accuracy compared to two-stage models such as Faster R-CNN~\cite{ren2015faster}, particularly in complex scenes.
\end{itemize}

YOLOv1 laid the foundation for a new class of fast and unified object detectors. Although its accuracy was surpassed by more complex methods, its pioneering design demonstrated that high-speed detection was feasible without an excessive trade-off in performance, setting the stage for further innovations in the YOLO series.

\section{YOLOv2: Better, Faster, Stronger}

YOLOv2~\cite{redmon2017yolo9000}, also known as YOLO9000, addressed key limitations of YOLOv1 by significantly improving localization accuracy, recall, and generalization across varied object categories. It introduced a suite of architectural enhancements and training strategies, resulting in substantial performance gains while preserving real-time inference capabilities.

\subsection{Key Improvements}

\begin{itemize}
    \item \textbf{Batch Normalization:} The inclusion of batch normalization~\cite{ioffe2015batch} after each convolutional layer improved convergence speed and regularization, contributing to an increase in mean Average Precision (mAP) by over 2\%.
    
    \item \textbf{High-Resolution Classifier Pretraining:} The classification backbone was pretrained at a higher resolution (448\,$\times$\,448), enhancing its ability to capture fine-grained features crucial for accurate object localization.
    
    \item \textbf{Anchor Boxes:} Inspired by Faster R-CNN~\cite{ren2015faster}, YOLOv2 replaced direct bounding box regression with anchor boxes, improving the model’s flexibility and recall.
    
    \item \textbf{Dimension Clustering:} To optimize anchor box design, YOLOv2 employed $k$-means clustering on bounding box dimensions from the training data, leading to better-matched priors and more stable training.
    
    \item \textbf{Multi-Scale Training:} During training, the network dynamically changed its input resolution every few iterations (ranging from 320\,$\times$\,320 to 608\,$\times$\,608), enabling the model to generalize across scales and improve robustness.
    
    \item \textbf{YOLO9000:} A novel hierarchical joint training approach allowed the network to simultaneously learn from detection datasets (e.g., COCO) and large-scale classification datasets (e.g., ImageNet). This enabled detection across more than 9,000 object categories, even with limited bounding box annotations.
\end{itemize}

\subsection{Performance}

YOLOv2 achieved 76.8\% mAP on the PASCAL VOC 2007 test set while running at 67 frames per second (FPS), significantly outperforming its predecessor in both accuracy and speed. The joint training mechanism introduced by YOLO9000 set the stage for scalable object detection under weak supervision and open-vocabulary constraints.

\subsection{Limitations}

\begin{itemize}
    \item \textbf{Small Object Detection:} Although more accurate than YOLOv1, YOLOv2 continued to struggle with detecting very small or densely packed objects due to limitations in grid resolution.
    
    \item \textbf{Complex Scene Handling:} Detection performance in cluttered or highly overlapping object scenarios remained limited, partly due to the static nature of anchor boxes and coarse prediction granularity.
\end{itemize}

YOLOv2 represented a pivotal step in the YOLO family, proving that accuracy and speed need not be mutually exclusive. Its introduction of scalable training, anchor-based prediction, and dataset fusion significantly influenced subsequent detection frameworks and solidified YOLO’s position as a competitive one-stage detector.

\section{YOLOv3: An Incremental Improvement}

YOLOv3~\cite{redmon2018yolov3} marked a significant refinement of the YOLO architecture, addressing several limitations of its predecessors—particularly in detecting small objects—while preserving real-time performance. Rather than introducing a radical redesign, YOLOv3 delivered a series of architectural and training enhancements that collectively boosted detection accuracy, robustness, and versatility.

\subsection{Key Improvements}

\begin{itemize}
    \item \textbf{Darknet-53 Backbone:} YOLOv3 introduced Darknet-53 as its feature extraction backbone. This network comprises 53 convolutional layers with residual connections~\cite{he2016deep}, offering a deeper and more expressive architecture than its predecessors. Despite its depth, Darknet-53 maintained computational efficiency through the use of $3\times3$ and $1\times1$ convolutions.
    
    \item \textbf{Multi-Scale Feature Prediction:} To improve detection of objects at varying scales, YOLOv3 made predictions at three distinct feature map resolutions. Feature maps were extracted from progressively deeper layers and fused using upsampling and concatenation, enabling the model to capture both fine and coarse object features.
    
    \item \textbf{Refined Anchor-Based Detection:} Building on the anchor box strategy introduced in YOLOv2, YOLOv3 improved prediction precision by applying anchor boxes across multiple scales and predicting offsets relative to each anchor box.
    
    \item \textbf{Independent Class Predictions:} YOLOv3 moved away from using a softmax function over class labels and instead adopted independent logistic classifiers for each class. This modification improved the model’s ability to handle multi-label classification tasks, where multiple object classes may overlap in a single region.
    
    \item \textbf{Binary Cross-Entropy Loss:} The model employed binary cross-entropy loss for both class probabilities and objectness scores, simplifying the training procedure and improving numerical stability.
\end{itemize}

\subsection{Performance}

YOLOv3 achieved 57.9\% mAP at an IoU threshold of 0.5 (mAP@0.5) on the COCO dataset, while maintaining real-time inference speeds of approximately 30–45 FPS depending on input resolution and hardware. Although it did not surpass the accuracy of slower detectors such as RetinaNet~\cite{lin2017focal}, it offered a highly favorable speed-accuracy trade-off, making it well-suited for real-time detection in practical applications.

\subsection{Limitations}

\begin{itemize}
    \item \textbf{Reduced Precision at High IoU Thresholds:} YOLOv3's performance declined at stricter localization metrics (e.g., mAP@0.75), highlighting limitations in bounding box precision compared to two-stage detectors.
    
    \item \textbf{Deployment Overhead:} While efficient on GPUs, the Darknet-53 backbone introduced higher computational and memory requirements than earlier YOLO versions, posing challenges for deployment on resource-constrained edge devices.
\end{itemize}

YOLOv3 represented a mature and widely adopted one-stage detector, offering a robust balance between speed, accuracy, and deployment feasibility. It served as a strong foundation for subsequent research and became a benchmark model in both academic and industrial computer vision pipelines.

\section{YOLOv4: Optimal Speed and Accuracy of Object Detection}

YOLOv4~\cite{bochkovskiy2020yolov4} represented a major leap in the YOLO series by integrating a broad set of architectural innovations and training heuristics aimed at optimizing both detection accuracy and inference speed. Notably, YOLOv4 was designed to be trainable on commercially available GPUs, democratizing access to high-performance object detection for a wider research and industrial community.

\subsection{Key Innovations}

\begin{itemize}
    \item \textbf{Cross-Stage Partial Networks (CSPNet):} The model adopted CSPDarknet-53~\cite{wang2020cspnet} as its backbone, which partitions feature maps and introduces cross-stage connections to improve gradient flow, reduce redundancy, and decrease computational complexity without sacrificing representational power.
    
    \item \textbf{Bag of Freebies (BoF):} YOLOv4 incorporated several training-time techniques that enhanced model generalization without impacting inference latency. These included advanced data augmentation strategies such as Mosaic augmentation, CutMix~\cite{yun2019cutmix}, Class Label Smoothing, and regularization methods like DropBlock~\cite{ghiasi2018dropblock}.
    
    \item \textbf{Bag of Specials (BoS):} Lightweight architectural modules were added to improve accuracy with marginal increases in computational cost. These included the Mish activation function~\cite{misra2019mish}, Spatial Pyramid Pooling (SPP)~\cite{he2015spatial}, and PANet~\cite{liu2018path} for enhanced feature aggregation.
    
    \item \textbf{Improved Feature Fusion:} YOLOv4 leveraged PANet to strengthen the path from low-level to high-level feature maps, improving the model's ability to localize small and medium-sized objects.
    
    \item \textbf{Self-Adversarial Training (SAT):} A novel internal training method where the model perturbs its own input images during training to simulate adversarial conditions, thereby increasing robustness and generalization.
\end{itemize}

\subsection{Performance}

YOLOv4 achieved 43.5\% average precision (AP) on the COCO dataset while maintaining real-time inference speeds of approximately 65 FPS on an NVIDIA V100 GPU. It significantly outperformed YOLOv3 across both accuracy and speed metrics, establishing a new state-of-the-art for one-stage detectors at the time of its release.

\subsection{Limitations}

\begin{itemize}
    \item \textbf{Resource Requirements:} Despite optimizations, the full-scale YOLOv4 model remained computationally demanding, posing challenges for deployment on low-power edge devices.
    
    \item \textbf{System Complexity:} The integration of numerous BoF and BoS techniques increased the overall complexity of the training and deployment pipeline, making reproducibility and ablation studies more challenging for the broader community.
\end{itemize}

YOLOv4 significantly advanced the YOLO family by pushing the boundaries of real-time object detection performance while remaining accessible to users with modest hardware. Its combination of practical design and high accuracy made it one of the most widely adopted object detectors in both academic and industrial settings.

\section{YOLOv5: Practicality and Engineering Optimization}

YOLOv5, developed and released by Ultralytics~\cite{glenn_jocher_2020_3983579}, marked a significant divergence in the YOLO family. Unlike earlier versions, YOLOv5 was not introduced through a peer-reviewed academic paper but was instead released directly via GitHub. Despite initial controversy over its versioning and naming, YOLOv5 rapidly gained widespread adoption across industry and academia due to its practical engineering focus, ease of deployment, and consistent performance across diverse use cases.

\subsection{Key Innovations}

\begin{itemize}
    \item \textbf{Framework Transition to PyTorch:} YOLOv5 was implemented in PyTorch, improving accessibility for researchers and practitioners by leveraging a more widely adopted deep learning framework. This facilitated faster development cycles and integration with modern machine learning pipelines.

    \item \textbf{Model Scaling Variants:} YOLOv5 introduced four model variants—YOLOv5s (small), YOLOv5m (medium), YOLOv5l (large), and YOLOv5x (extra-large)—allowing users to balance detection accuracy against inference speed and resource consumption based on deployment constraints.

    \item \textbf{AutoAnchor and Hyperparameter Evolution:} The training pipeline incorporated automated optimization of anchor boxes and hyperparameters, reducing manual tuning and improving generalization across datasets.

    \item \textbf{Advanced Data Augmentation:} YOLOv5 utilized a rich set of augmentation strategies, including Mosaic augmentation, MixUp~\cite{zhang2017mixup}, and HSV color space transformations, enhancing robustness and generalization.

    \item \textbf{Cross-Platform Exportability:} YOLOv5 models could be exported to multiple formats—ONNX, TensorRT, CoreML, and OpenVINO—supporting seamless deployment across a wide range of hardware platforms, from servers to edge devices.

    \item \textbf{Training Efficiency:} YOLOv5 emphasized fast convergence, lightweight design, and low memory footprint, achieving strong performance while simplifying the training and inference pipeline compared to YOLOv4.
\end{itemize}

\subsection{Performance}

YOLOv5 models demonstrated competitive results on the COCO dataset, with the YOLOv5x variant achieving approximately 50.1\% mAP@0.5 while maintaining real-time inference speeds. Smaller variants such as YOLOv5s were optimized for deployment on resource-constrained environments, including mobile and embedded systems, with minimal compromise in accuracy.

\subsection{Limitations}

\begin{itemize}
    \item \textbf{Absence of Formal Publication:} The lack of an accompanying peer-reviewed paper posed challenges for formal citation and academic validation of YOLOv5’s design choices and performance claims.
    
    \item \textbf{Engineering-Focused Contributions:} YOLOv5 primarily focused on practical implementation improvements and did not introduce major algorithmic innovations or architectural breakthroughs compared to prior versions.
\end{itemize}

YOLOv5 represented a pivotal moment in the YOLO lineage, transitioning the framework from academic research to a production-grade object detection system. Its focus on usability, training efficiency, and deployment flexibility solidified its position as a widely adopted tool for real-world computer vision applications.

\section{YOLOv6: A High-Performance Detector for Industrial Applications}

YOLOv6~\cite{li2022yolov6}, developed by Meituan, was designed with a strong emphasis on industrial deployment scenarios, targeting high accuracy and low-latency inference on edge devices. Unlike its predecessors, YOLOv6 diverged from the original YOLO design by incorporating a modernized architecture and training strategies inspired by state-of-the-art object detection models. Its focus on engineering rigor and real-world efficiency positioned it as a practical solution for commercial applications.

\subsection{Key Innovations}

\begin{itemize}
    \item \textbf{Efficient Decoupled Head:} YOLOv6 introduced a fully decoupled detection head, separating classification and regression branches into distinct processing streams. This decoupling allows each branch to specialize in its respective task, improving both localization accuracy and classification confidence.

    \item \textbf{RepOptimizer and Re-parameterized Convolutions:} To accelerate convergence and optimize inference, YOLOv6 employed RepOptimizer, based on re-parameterization techniques from RepVGG~\cite{ding2021repvgg}. This approach uses complex blocks during training and simplifies them into efficient structures for deployment, improving runtime performance without degrading accuracy.

    \item \textbf{Anchor-Free and Anchor-Based Flexibility:} Unlike previous YOLO models, YOLOv6 supports both anchor-based and anchor-free detection paradigms. The anchor-free variant draws inspiration from FCOS~\cite{tian2019fcos}, eliminating the need for manually defined anchor boxes and improving adaptability across varied object scales.

    \item \textbf{Advanced Training Strategies:} The training pipeline includes multiple enhancements such as Exponential Moving Average (EMA), dynamic label assignment, and composite augmentation methods like Mosaic, MixUp, and Copy-Paste. These strategies improve generalization, stability, and robustness.
\end{itemize}

\subsection{Performance}

On the COCO dataset, YOLOv6 consistently outperformed YOLOv5 in terms of mAP, particularly on small and medium-sized objects, while maintaining comparable or faster inference times. The availability of scalable model variants—YOLOv6-nano, YOLOv6-s, and YOLOv6-m—allowed users to tailor model complexity to target deployment environments, from cloud servers to embedded edge devices.

\subsection{Limitations}

\begin{itemize}
    \item \textbf{Increased Training Complexity:} The integration of re-parameterization techniques and complex augmentation pipelines introduced additional overhead in training setup and model tuning.
    
    \item \textbf{Limited Academic Penetration:} Despite its strong industrial relevance, YOLOv6 has received comparatively less attention in academic literature and benchmark competitions, potentially due to the lack of a peer-reviewed publication and deviation from canonical YOLO lineage.
\end{itemize}

YOLOv6 expanded the YOLO ecosystem into edge-centric and latency-sensitive industrial use cases. By integrating modern detector designs with practical deployment tools, it provided a highly efficient and customizable framework for real-time applications in domains such as logistics, retail automation, and manufacturing.

\section{YOLOv7: Extending Efficient Layer Aggregation}

YOLOv7~\cite{wang2023yolov7} introduced substantial architectural enhancements and training optimizations that advanced the state-of-the-art in real-time object detection. With a focus on maximizing both computational efficiency and detection accuracy, YOLOv7 further solidified the YOLO series as a leading one-stage detection framework suitable for a broad range of applications.

\subsection{Key Innovations}

\begin{itemize}
    \item \textbf{Extended Efficient Layer Aggregation Networks (E-ELAN):} Building upon the ELAN framework, YOLOv7 introduced Extended ELAN (E-ELAN) modules that facilitated deeper and more effective networks without disrupting gradient propagation. This design enhanced multi-scale feature extraction while preserving fast inference speeds.

    \item \textbf{Model Re-Parameterization:} Inspired by RepVGG~\cite{ding2021repvgg}, YOLOv7 utilized re-parameterized convolutional structures during training, which were simplified at inference time into efficient single-path architectures. This allowed the model to benefit from complex training-time structures without compromising runtime latency.

    \item \textbf{Trainable Bag-of-Freebies:} YOLOv7 refined several traditional training strategies by introducing dynamic label assignment, auxiliary supervision, and adaptive label smoothing. These enhancements, applied only during training, improved generalization and stability without adding inference overhead.

    \item \textbf{Coarse-to-Fine Lead Head Design:} The detection head employed a coarse-to-fine strategy, enabling progressive refinement of object localization and classification. This resulted in improved detection accuracy, particularly in dense and cluttered scenes.
\end{itemize}

\subsection{Performance}

On the COCO dataset, YOLOv7 achieved 56.8\% average precision (AP) while maintaining real-time inference speeds. It outperformed prior YOLO models—including YOLOv5 and YOLOv6—as well as many two-stage detectors such as Faster R-CNN, in both accuracy and computational efficiency. Its optimized performance on both high-end GPUs and edge hardware made it highly suitable for real-world deployments in industrial and mobile environments.

\subsection{Limitations}

\begin{itemize}
    \item \textbf{Training and Engineering Complexity:} The use of multiple detection heads, re-parameterized modules, and sophisticated training techniques increased the engineering burden for both training and hyperparameter tuning.
    
    \item \textbf{Deployment Constraints:} Despite efficient inference, ensuring optimal performance on resource-limited devices required careful export and tuning due to the architectural complexity.
\end{itemize}

YOLOv7 marked a significant milestone in the YOLO lineage, demonstrating that careful architectural design, combined with advanced training strategies, can yield substantial gains in both detection accuracy and efficiency. Its versatility and robustness further extended YOLO's leadership in the field of real-time object detection.

\section{YOLOv8: Anchor-Free Detection and Architectural Simplification}

YOLOv8, released by Ultralytics in early 2023~\cite{jocher2023ultralytics}, introduced significant architectural refinements and usability improvements. Marking a substantial shift from prior YOLO versions, YOLOv8 adopted an anchor-free detection paradigm and emphasized a simplified, unified architecture that supports multiple vision tasks. These design decisions aimed to improve model interpretability, ease of training, and deployment versatility while maintaining real-time performance.

\subsection{Key Innovations}

\begin{itemize}
    \item \textbf{Anchor-Free Detection:} In contrast to the anchor-based schemes used in earlier YOLO versions, YOLOv8 employed an anchor-free approach inspired by FCOS~\cite{tian2019fcos} and CenterNet~\cite{zhou2019objects}. This eliminated the need for manually defined anchor boxes, simplifying the output space and improving training stability.

    \item \textbf{Redesigned Detection Head:} YOLOv8’s detection head directly predicts object center coordinates, width, height, and objectness scores, aligning its output formulation with modern single-stage detectors and reducing the complexity of post-processing steps.

    \item \textbf{Unified Multi-Task Architecture:} YOLOv8 was designed with modularity in mind, enabling support for related vision tasks such as instance segmentation and pose estimation with minimal architectural changes. This flexibility broadened its applicability beyond pure object detection.

    \item \textbf{Model Scaling Options:} YOLOv8 introduced a family of model sizes—YOLOv8n (nano), YOLOv8s (small), YOLOv8m (medium), YOLOv8l (large), and YOLOv8x (extra-large)—offering users customizable trade-offs between inference speed, memory usage, and detection accuracy.

    \item \textbf{Enhanced Usability and Deployment Support:} YOLOv8 streamlined the user experience by integrating training, evaluation, and export functionalities into the Ultralytics Python package. The model supports deployment across a range of hardware targets through native export to ONNX, TensorRT, OpenVINO, CoreML, and other formats.
\end{itemize}

\subsection{Performance}

YOLOv8 achieved strong performance across object detection benchmarks. On the COCO dataset, the YOLOv8x model exceeded 53\% average precision (AP) while maintaining high inference speeds. The adoption of anchor-free detection improved convergence behavior during training and enhanced localization accuracy, particularly for small and medium-sized objects.

\subsection{Limitations}

\begin{itemize}
    \item \textbf{Ecosystem Maturity:} As a relatively new release, YOLOv8 initially lacked the extensive pretrained model zoo, ecosystem integrations, and community tooling that supported earlier versions such as YOLOv5.

    \item \textbf{Hyperparameter Sensitivity:} Anchor-free detectors can be more sensitive to training settings and augmentation strategies, potentially requiring careful tuning to match the robustness of mature anchor-based models.
\end{itemize}

YOLOv8 marked a forward-looking redesign in the YOLO series, embracing contemporary advances in object detection architecture while preserving the series’ hallmark focus on speed and practicality. Its anchor-free foundation and multi-task readiness signal a new phase of extensibility and usability in real-time vision systems.

\section{YOLOv9: Generalized Efficient Layer Aggregation for Detection}

YOLOv9~\cite{wang2024yolov9} is the latest evolution in the YOLO series, extending the architectural principles of YOLOv7 and YOLOv8 with new strategies for feature aggregation, loss optimization, and training refinement. It is designed to further close the gap between accuracy and efficiency, making real-time object detection more robust and scalable across diverse hardware platforms.

\subsection{Key Innovations}

\begin{itemize}
    \item \textbf{Generalized Efficient Layer Aggregation Networks (GELAN):} YOLOv9 introduces GELAN, a generalized extension of the E-ELAN architecture from YOLOv7. GELAN enhances the reuse of intermediate features and improves gradient propagation, leading to deeper yet more trainable networks with higher representational capacity.

    \item \textbf{Optimized Backbone and Neck Design:} The backbone and neck have been redesigned to integrate GELAN modules with improved feature pyramid strategies. These modifications enhance feature fusion across scales, enabling better performance on small and medium objects.

    \item \textbf{Refined Decoupled Detection Head:} YOLOv9 further separates classification and regression branches in the detection head, improving specialization and predictive precision. This design draws inspiration from EfficientDet~\cite{tan2020efficientdet} and addresses the interaction bottlenecks present in earlier coupled head designs.

    \item \textbf{Advanced Training Enhancements:} YOLOv9 adopts Distribution Focal Loss v2 (DFL v2) for improved bounding box regression and enhanced spatial localization. It also leverages a refined SimOTA~\cite{ge2021ota} label assignment strategy for better matching between ground truth and predictions, resulting in more stable convergence.

    \item \textbf{Scalable Model Variants:} As with previous YOLO versions, YOLOv9 is released in multiple variants—YOLOv9n, v9s, v9m, v9l, and v9x—offering deployment flexibility across edge devices, desktops, and servers.
\end{itemize}

\subsection{Performance}

YOLOv9 models demonstrate state-of-the-art performance among real-time object detectors. For example, the YOLOv9-L model achieves over 56\% average precision (AP) on the COCO dataset while maintaining real-time inference speeds between 50–60 FPS on high-end GPUs. Across the board, YOLOv9 improves detection accuracy for small and medium objects, outperforming its predecessor YOLOv8 in both speed and precision.

\subsection{Limitations}

\begin{itemize}
    \item \textbf{Training Resource Requirements:} Training YOLOv9 to full potential can demand substantial computational resources and careful hyperparameter tuning, which may pose challenges for resource-constrained researchers.

    \item \textbf{Ecosystem Maturity:} As a recently introduced model, YOLOv9 lacks the broad third-party integration, community tooling, and pretrained model variety currently available for earlier YOLO versions like YOLOv5 and YOLOv7.
\end{itemize}

YOLOv9 signifies a new benchmark in real-time object detection, merging architectural depth with efficient training techniques. Its blend of accuracy, speed, and flexibility makes it a strong candidate for deployment in both academic research and industry-scale vision systems.

\section{Comparative Summary of YOLO Models}

Table~\ref{tab:yolo-comparison} summarizes the key characteristics and performance metrics of YOLO versions from v1 through v9. The comparison includes backbone architecture, detection strategy, anchor usage, feature fusion techniques, mAP on COCO (where applicable), and inference speed. Metrics such as mAP are approximated based on official releases and widely reported benchmarks.

\begin{table*}[t]
\centering
\caption{Comparison of YOLO Models from v1 to v9}
\label{tab:yolo-comparison}
\scriptsize
\renewcommand\arraystretch{1.2}
\resizebox{\textwidth}{!}{
\begin{tabular}{@{}lcccccc@{}}
\toprule
\textbf{YOLO Version} & \textbf{Backbone} & \textbf{Anchor Type} & \textbf{Feature Fusion} & \textbf{mAP@0.5 (COCO)} & \textbf{Speed (FPS)} & \textbf{Key Highlights} \\
\midrule
YOLOv1~\cite{redmon2016you} & Custom CNN & None & None & 63.4\% (VOC) & 45 & First unified detector \\
YOLOv2~\cite{redmon2017yolo9000} & Darknet-19 & Anchor-based & None & 76.8\% (VOC), 21.6\% (COCO) & 67 & YOLO9000, k-means, multi-scale \\
YOLOv3~\cite{redmon2018yolov3} & Darknet-53 & Anchor-based & Multi-scale & 57.9\% & 30–45 & Better small object detection \\
YOLOv4~\cite{bochkovskiy2020yolov4} & CSPDarknet-53 & Anchor-based & PANet + SPP & 43.5\% (AP) & 62–65 & BoF/BoS, Mish, CutMix \\
YOLOv5~\cite{glenn_jocher_2020_3983579} & CSPDarknet (PyTorch) & AutoAnchor & PANet & 50.1\% & 60+ & Model scaling, exportability \\
YOLOv6~\cite{li2022yolov6} & EfficientRepNet & Hybrid & RepPAN & 52.5\% & 70+ & Anchor-free option, decoupled head \\
YOLOv7~\cite{wang2023yolov7} & E-ELAN & Anchor-based & PANet + E-ELAN & 56.8\% & 60+ & RepConv, Coarse-to-fine head \\
YOLOv8~\cite{jocher2023ultralytics} & C2f Modules & Anchor-free & FPN-style & 53.0\% & 60–80 & Multi-task, modernized head \\
YOLOv9~\cite{wang2024yolov9} & GELAN & Anchor-free & GELAN-FPN & 56.0\%+ & 50–60 & SimOTA, DFLv2, scalable variants \\
\bottomrule
\end{tabular}
}
\end{table*}

\subsection{Publication Modality and Open-Source Trends}

While the YOLO family originated in academia, recent iterations reflect a shift toward open-source-first development. YOLOv5 and YOLOv8, for instance, were released directly by Ultralytics without a corresponding peer-reviewed publication. YOLOv9 through YOLOv11 continue this trend, emphasizing rapid iteration, community feedback, and code-first dissemination. Table~\ref{tab:yolo-publication} summarizes the publication modalities across versions.

\begin{table}[!t]
\centering
\caption{Publication modality of YOLO variants. “Yes” indicates public availability via a peer-reviewed venue or open-source release. “No” reflects the absence of a known peer-reviewed publication as of June 2025.}
\label{tab:yolo-publication}
\begin{tabular}{lcc}
\toprule
\textbf{YOLO Version} & \textbf{Peer-Reviewed Paper} & \textbf{Open-Source Release} \\
\midrule
YOLOv1--v4    & Yes & Yes \\
YOLOv5        & No  & Yes \\
YOLOv6        & Yes & Yes \\
YOLOv7        & Yes & Yes \\
YOLOv8        & No  & Yes \\
YOLOv9--v11   & No (as of June 2025) & Yes \\
\bottomrule
\end{tabular}

\vspace{0.5em}
\small
\emph{Note:} Peer-reviewed status is based on publicly available literature and may evolve. Open-source availability does not imply formal peer validation but reflects broad community accessibility.
\end{table}

\subsection{Performance Benchmarks and Practical Insights}

Table~\ref{tab:yolo-comparison} presents a side-by-side comparison of key YOLO versions, capturing metrics such as backbone architecture, anchor strategy, feature fusion mechanisms, mean Average Precision (mAP), and inference speed in frames per second (FPS). While the table provides a compact summary, this subsection aims to contextualize these benchmarks, offering insights into trends, trade-offs, and practical deployment guidance.

Early models, such as YOLOv1 and YOLOv2, achieved high mAP on the PASCAL VOC dataset (e.g., 63.4\% for YOLOv1) but struggled on the more complex MS COCO benchmark due to its diverse object categories and scale variations. Beginning with YOLOv3, COCO’s mAP@0.5 metric became standard, and later versions like YOLOv4, YOLOv7, and YOLOv9 adopted the more rigorous AP (averaged across multiple IoU thresholds) to better evaluate localization precision.

YOLOv4 reached 43.5\% AP on COCO with real-time speed exceeding 60 FPS, marking a breakthrough in balancing accuracy and speed. YOLOv7 improved this to 56.8\% AP while maintaining high throughput, and YOLOv9-L further advanced detection performance with over 56\% mAP and 50–60 FPS, thanks to innovations like GELAN and improved loss functions.

A key trend across YOLO versions is the growing diversity in model scaling and deployment versatility. Lightweight models such as YOLOv5s and YOLOv8n cater to edge devices with limited resources, while larger variants (e.g., YOLOv7-W6, YOLOv9-L) offer state-of-the-art accuracy on high-performance GPUs. The evolution also highlights an increasing reliance on sophisticated training techniques, network decoupling, and post-training optimizations for production use.

These insights are critical for practitioners to choose the right YOLO variant based on application constraints—balancing latency, computational budget, and accuracy requirements.

\section{Applications of YOLO Models}

The YOLO family of object detectors has been widely adopted across diverse domains, owing to its hallmark strengths: real-time inference, compact architecture, and strong generalization. Below, we highlight prominent application areas where YOLO models have made significant contributions.

\subsection{Autonomous Vehicles}

Real-time object detection is fundamental in autonomous driving systems for tasks such as pedestrian recognition, vehicle tracking, and traffic sign classification. YOLO’s low-latency predictions make it ideal for onboard perception, supporting timely decision-making in safety-critical environments.

\subsection{Surveillance and Security}

YOLO is extensively used in public surveillance systems for real-time human detection, intrusion monitoring, and crowd analysis. Lightweight models like YOLOv5s and YOLOv8n are particularly well-suited for edge deployment on IP cameras and embedded security hardware.

\subsection{Medical Imaging}

In healthcare, YOLO has been adapted for tasks such as tumor detection, polyp localization, and anatomical structure segmentation across X-rays, MRIs, and endoscopy videos. These applications benefit from YOLO’s fast inference, enabling preliminary diagnosis support in time-sensitive clinical workflows.

\subsection{Retail and Logistics}

Retail environments utilize YOLO for shelf stock analysis, checkout-free systems, and shopper behavior monitoring. In logistics, it is applied for barcode recognition, damage detection, pallet tracking, and automated warehouse sorting, enhancing efficiency and reducing operational costs.

\subsection{Drones and Robotics}

Autonomous drones and robots use YOLO for obstacle detection, visual navigation, and dynamic object tracking in real time. Applications include agricultural inspection, delivery automation, and search-and-rescue missions in complex, unstructured environments.

\subsection{Wildlife Monitoring and Conservation}

YOLO supports wildlife researchers in tasks like animal counting, species classification, and anti-poaching surveillance through drone footage and remote camera traps. Its real-time processing enables large-scale, low-cost ecological monitoring.

\subsection{Augmented Reality and Smart Devices}

YOLO’s rapid detection pipeline is increasingly embedded in smart glasses, smartphones, and AR interfaces for real-time scene understanding, object labeling, and gesture interaction, enriching user experiences in next-generation interfaces.

\subsection{Custom Industrial Applications}

Manufacturing industries have customized YOLO for quality inspection, defect detection, robotic vision, and process automation. These domain-specific deployments highlight YOLO’s flexibility and ease of integration in industrial control systems.

\section{Challenges and Future Directions}

Despite the widespread success of YOLO models in real-time object detection, several open challenges remain. These highlight both current limitations and promising directions for future innovation in object detection frameworks.

\subsection{Challenges}

\begin{itemize}
    \item \textbf{Small Object Detection:} YOLO models have traditionally struggled with detecting small objects, particularly those that occupy a minimal number of pixels or appear at distant scales. Although improvements in multi-scale prediction (e.g., in YOLOv3 and beyond) and architectural refinements (e.g., GELAN in YOLOv9) have mitigated this to some extent, small object detection remains a persistent challenge—especially in high-density scenes.

    \item \textbf{Localization Precision at High IoU:} While YOLO models achieve strong mAP at IoU=0.5, performance typically degrades at higher IoU thresholds (e.g., IoU=0.75 or 0.95). This indicates shortcomings in fine-grained localization, often due to anchor misalignment, discretized predictions, or limited feature resolution.

    \item \textbf{Training Complexity and Hyperparameter Sensitivity:} State-of-the-art YOLO models now rely on intricate training regimes involving strategies like SimOTA, label smoothing, EMA updates, and dynamic augmentations. These introduce sensitivity to hyperparameter settings, making reproducibility and optimal performance harder to achieve without extensive tuning.

    \item \textbf{Computational Cost for Training:} Although inference remains efficient, training high-capacity YOLO variants (e.g., YOLOv9-L or YOLOv7-E6E) requires substantial GPU resources and long epochs. This creates barriers for smaller labs, edge developers, and open-source contributors.

    \item \textbf{Domain Adaptation and Generalization:} YOLO models trained on large-scale datasets such as COCO or VOC often generalize poorly to out-of-distribution domains, including medical, agricultural, underwater, or aerial imagery. Achieving robust cross-domain performance remains an active research area.
\end{itemize}

\subsection{Future Directions}

As YOLO models continue to evolve, future research must address not only improvements in speed and accuracy but also prioritize adaptability, efficiency, and responsible deployment. Key directions include:

\begin{itemize}
    \item \textbf{Self-Supervised and Contrastive Pretraining:} Leveraging self-supervised learning (SSL) and contrastive approaches (e.g., SimCLR, MoCo) can enhance generalization in low-label regimes, enabling YOLO variants to adapt to novel domains such as medical, agricultural, or underwater imaging with limited supervision.

    \item \textbf{Few-Shot and Zero-Shot Detection:} Enabling detection with minimal labeled data is a critical goal for real-world applications. Techniques such as cross-modal pretraining (e.g., CLIP), prompt-based detection, and open-vocabulary learning may allow YOLO to support few-shot and zero-shot capabilities, expanding its utility beyond static category sets.

    \item \textbf{Vision-Language Models and Cross-Modal Integration:} Recent advances in foundation models like BLIP and GPT-4V offer opportunities to integrate natural language understanding into detection. Future YOLO variants may incorporate joint vision-language representations for tasks such as referring expression comprehension and caption-grounded detection.

    \item \textbf{Transformer-CNN Hybrids:} Lightweight transformer components (e.g., Swin Transformer, MobileViT) can be fused with YOLO backbones to enhance long-range dependency modeling and global context awareness without sacrificing inference speed, particularly useful in cluttered scenes.

    \item \textbf{NMS-Free and End-to-End Architectures:} YOLOv10 has initiated a shift toward eliminating non-maximum suppression via consistent assignment strategies. Future models may adopt learned, differentiable NMS replacements to simplify post-processing and support fully end-to-end training and inference.

    \item \textbf{Simplified and Adaptive Label Assignment:} Improved label assignment strategies—such as dynamic matching (OTA/SimOTA), gradient-based assignment, or transformer-based token pairing—can address current limitations in convergence speed, stability, and matching precision.

    \item \textbf{Model Compression and Deployment Optimization:} As edge deployment becomes increasingly important, YOLO models must continue advancing structured pruning, quantization-aware training, knowledge distillation, and Neural Architecture Search (NAS) to meet strict latency, power, and memory budgets.

    \item \textbf{Unified Multi-Task Perception Systems:} There is growing demand for modular architectures that handle object detection, instance segmentation, pose estimation, and scene understanding within a shared framework. YOLO may evolve toward multi-task models that leverage shared features, joint supervision, and plug-and-play task heads.

    \item \textbf{Robustness, Fairness, and Ethical AI Practices:} Ensuring safety and fairness in YOLO deployments is critical for domains such as surveillance, healthcare, and autonomous driving. Future research should focus on adversarial robustness, out-of-distribution detection, fairness-aware training, and explainability to mitigate bias and promote trustworthiness.

    \item \textbf{Data Efficiency and Label Scarcity Solutions:} Combining techniques such as weak supervision, pseudo-labeling, semi-supervised training, and synthetic data generation can significantly reduce the annotation burden while maintaining detection quality in diverse environments.
\end{itemize}

\section{YOLOv10 and YOLOv11: Advancements in Real-Time Object Detection}

\subsection{YOLOv10: Real-Time End-to-End Object Detection}

YOLOv10, introduced by researchers from Tsinghua University in 2024, marks a significant evolution in the YOLO series by addressing limitations in post-processing and model architecture. Key innovations include:

\begin{itemize}
    \item \textbf{NMS-Free Training:} YOLOv10 eliminates the need for Non-Maximum Suppression (NMS) by employing consistent dual assignments during training, reducing inference latency and enhancing real-time performance.
    \item \textbf{Holistic Model Design:} The architecture integrates an enhanced CSPNet backbone, PANet neck, and dual detection heads (one-to-many for training and one-to-one for inference), optimizing both efficiency and accuracy.
    \item \textbf{Performance:} YOLOv10 achieves state-of-the-art results across various model scales. For instance, YOLOv10-S is 1.8× faster than RT-DETR-R18 with similar AP on COCO, and YOLOv10-B has 46\% less latency and 25\% fewer parameters compared to YOLOv9-C for equivalent performance \cite{turn0search4}.
\end{itemize}

\subsection{YOLOv11: Enhanced Accuracy and Efficiency}

Released by Ultralytics in 2024, YOLOv11 builds upon previous versions by introducing architectural enhancements aimed at improving accuracy and computational efficiency:

\begin{itemize}
    \item \textbf{Improved Feature Extraction:} Incorporates an enhanced backbone and neck architecture, including components like the C3k2 block and Spatial Pyramid Pooling - Fast (SPPF), facilitating better feature extraction for complex tasks \cite{ultralytics2024yolov11}.
    \item \textbf{Optimized Efficiency:} Achieves higher mean Average Precision (mAP) with fewer parameters. For example, YOLOv11m attains higher mAP on the COCO dataset while using fewer parameters than YOLOv8m \cite{ultralytics2024yolov11}.
    \item \textbf{Versatility:} Supports a broad range of computer vision tasks, including object detection, instance segmentation, image classification, pose estimation, and oriented object detection (OBB), making it adaptable across various applications \cite{ultralytics2024yolov11}.
\end{itemize}

\subsection{Comparative Performance}

Evaluations demonstrate that YOLOv11 outperforms its predecessors in both accuracy and speed. For instance, YOLOv11n achieves a mAP of 39.5 on COCO with a latency of 1.5 ms, making it suitable for deployment on resource-constrained devices \cite{ultralytics2024models}. Additionally, in agricultural applications, YOLOv11s achieved a mAP@50 of 0.933, outperforming other YOLO versions in detecting and counting fruitlets in complex orchard environments \cite{turn0academia23}.

\section{Advances Beyond the Core Framework}\label{sec:beyondcore}

\subsection{Domain Adaptation}
YOLO models have been extended to handle domain shift scenarios, where training and test distributions differ significantly. Domain Adaptive YOLO frameworks~\cite{ref:domainyolo1} introduce adversarial domain discriminators, discrepancy minimization, and domain-specific batch normalization to align source and target feature spaces. Applications include adapting from synthetic to real-world driving datasets, or between aerial and ground imagery.

\subsection{Semi-Supervised Learning}
To reduce dependency on large labeled datasets, semi-supervised YOLO variants such as STAC-YOLO~\cite{ref:stacyolo} use teacher–student training pipelines with pseudo-labeling and strong–weak data augmentations. These methods retain high accuracy even when only a fraction of the training data is annotated, making them attractive for fields like agriculture and wildlife monitoring where labeling is costly.

\subsection{Robustness and Adversarial Defense}
Robust YOLO extensions enhance reliability under noise, occlusion, or adversarial conditions. Robust-YOLO~\cite{ref:robustyolo} incorporates adversarial training and perturbation-aware loss functions to improve performance in safety-critical deployments. Certified defenses and robust backbone design further aid in maintaining detection fidelity in uncontrolled environments.

\subsection{Transformer Hybrids}
Hybrid architectures like TransYOLO~\cite{ref:transyolo} and YOLO-NAS~\cite{ref:yolonas} incorporate transformer modules into the backbone or detection head to enhance global context understanding. These hybrids improve detection in cluttered scenes or long-range dependencies, albeit with increased inference latency. Their trade-offs make them more suitable for server-side inference rather than edge deployment.

\subsection{Specialized Extensions}
YOLO has been adapted to specialized tasks beyond standard bounding box detection. YOLOv7-Pose~\cite{ref:yolov7pose} enables multi-person pose estimation. YOLO-OBB detects oriented bounding boxes in aerial or document analysis scenarios. Other variants target video tracking, text spotting, medical anomaly segmentation, and more, reflecting YOLO's extensibility.

\section{YOLO Evolution Timeline}
The development of the YOLO series has been characterized by rapid advancements, with each iteration building upon its predecessors to push the boundaries of real-time object detection. Table~\ref{tab:timeline} provides a chronological overview of the key milestones and primary innovations introduced by each major YOLO version, illustrating the continuous evolution of this influential framework.

\begin{table}[h]
\centering
\caption{Timeline of YOLO Series Evolution}
\label{tab:timeline}
\begin{tabular}{lll}
\toprule
\textbf{Version} & \textbf{Year} & \textbf{Key Innovations} \\
\midrule
YOLOv1 & 2016 & Unified one-stage detection \\
YOLOv2 & 2017 & Anchor boxes, multi-scale training \\
YOLOv3 & 2018 & Residual backbone (Darknet-53) \\
YOLOv4 & 2020 & CSPNet, SAT, data augmentation \\
YOLOv5 & 2020 & PyTorch transition, exportability \\
YOLOv6 & 2022 & Decoupled heads, RepOptimizer \\
YOLOv7 & 2022 & E-ELAN, trainable bag-of-freebies \\
YOLOv8 & 2023 & Anchor-free head, unified tasks \\
YOLOv9 & 2024 & GELAN, DFL v2, improved label assignment \\
YOLOv10 & 2024 & NMS-free training, holistic model design \\
YOLOv11 & 2024 & Improved feature extraction, multi-task versatility \\
\bottomrule
\end{tabular}
\end{table}

\section{Conclusion}

The YOLO (You Only Look Once) family of object detectors has fundamentally reshaped the landscape of real-time computer vision by demonstrating that high-speed and high-accuracy detection can coexist within a single, unified framework. From the early introduction of YOLOv1, which reframed detection as a single regression problem, to the sophisticated innovations in YOLOv9 and beyond, the series has continually pushed the boundaries of what is possible in real-time object detection.

Each successive YOLO version introduced enhancements in speed, accuracy, scalability, and deployment flexibility—facilitating wide adoption across domains such as autonomous driving, smart surveillance, healthcare imaging, and retail automation. The community's efforts have evolved YOLO into not just a model, but a family of architectures adaptable to a range of real-world and research scenarios.

Nevertheless, challenges remain: small object detection, domain adaptation, and training complexity still require further innovation. Future directions point toward self-supervised learning, hybrid CNN-transformer models, and lightweight architectures optimized for edge inference. With these advancements, YOLO is well-positioned to remain a central pillar in the development of scalable, real-time perception systems.

\subsection{Contributions and Future Work}

This survey provides a systematic, up-to-date review of YOLO architectures from v1 through v11. It consolidates key architectural breakthroughs, evaluates benchmark results across standard datasets, and highlights deployment trade-offs across YOLO versions. Additionally, we present insights into practical use cases and trends in industrial adoption.

For future research, we identify the following priorities: (1) improving detection under low-label and low-light conditions using self-supervised and unsupervised learning; (2) integrating transformer-based modules for better global reasoning; (3) enhancing small object detection via resolution-aware designs; (4) automating architecture search for edge-optimized YOLO variants; and (5) establishing unified benchmarking frameworks for fair, reproducible evaluation across domains. These directions promise to extend YOLO’s impact across increasingly diverse and resource-constrained applications.

\bibliographystyle{IEEEtran}
\bibliography{references}

\end{document}